\documentclass{ecai} 
\usepackage{latexsym}
\usepackage{amssymb}
\usepackage{amsmath}
\usepackage{amsthm}
\usepackage{booktabs}
\usepackage{enumitem}
\usepackage{graphicx}
\usepackage{color}
\usepackage{hyperref}
\usepackage{algorithm}
\usepackage{algpseudocode}

\newcommand{\BibTeX}{B\kern-.05em{\sc i\kern-.025em b}\kern-.08em\TeX}

\begin{document}

%%%%%%%%%%%%%%%%%%%%%%%%%%%%%%%%%%%%%%%%%%%%%%%%%%%%%%%%%%%%%%%%%%%%%%%%

\begin{frontmatter}

%%% Use this command to specify your submission number.
%%% In doubleblind mode, it will be printed on the first page.

\paperid{1154} 

%%% Use this command to specify the title of your paper.

\title{Harnessing Structured Knowledge: A Concept Map-Based Approach for High-Quality Multiple Choice Question Generation with Effective Distractors}

% Structuring Knowledge through Hierarchical Concept Map Helps Language Models to Generate High Quality Multiple Choice Questions with Distractors
% Harnessing Structured Knowledge: A Concept Map-Based Approach for Generating Pedagogically Sound Multiple Choice Questions with Distractors

%%% Use this combinations of commands to specify all authors of your 
%%% paper. Use \fnms{} and \snm{} to indicate everyone's first names 
%%% and surname. This will help the publisher with indexing the 
%%% proceedings. Please use a reasonable approximation in case your 
%%% name does not neatly split into "first names" and "surname".
%%% Specifying your ORCID digital identifier is optional. 
%%% Use the \thanks{} command to indicate one or more corresponding 
%%% authors and their email address(es). If so desired, you can specify
%%% author contributions using the \footnote{} command.

\author[A]{\fnms{Nicy}~\snm{Scaria}\footnote{Equal contribution.}\thanks{Corresponding Author. Email: nicyscaria@iisc.ac.in}}
\author[A, B]{\fnms{Silvester}~\snm{John Joseph Kennedy}\footnotemark}
\author[A]{\fnms{Diksha}~\snm{Seth}}
\author[A]{\fnms{Ananya}~\snm{Thakur}} 
\author[A]{\fnms{Deepak}~\snm{Subramani}} 

\address[A]{Indian Institute of Science}
\address[B]{Talking Yak}

%%% Use this environment to include an abstract of your paper.

\begin{abstract}
Generating high-quality multiple choice questions (MCQs), especially those targeting diverse cognitive levels and incorporating common misconceptions into distractor design, is time-consuming and expertise-intensive, making manual creation impractical at scale. Current automated approaches typically generate questions at lower cognitive levels and fail to incorporate domain-specific misconceptions as distractors. This paper presents a hierarchical concept map-based framework that provides structured knowledge to guide Large Language Models (LLMs) in generating high-quality MCQs with distractors. We chose high-school physics as our test domain and began by developing a comprehensive hierarchical concept map covering major Physics topics and their interconnections with an efficient database design. Next, through an automated pipeline, topic-relevant sections of these concept maps are retrieved to serve as a structured context for the LLM to generate questions and distractors that specifically target common misconceptions. Lastly, an automated validation is completed to ensure that the generated MCQs meet the requirements provided. We evaluate our framework against two baseline approaches: a base LLM and a Retrieval Augmented Generation (RAG) based generation. We conducted expert evaluations and student assessments of the generated MCQs. Expert evaluation shows that our method significantly outperforms the baseline approaches, achieving a success rate of 75.20\% in meeting all quality criteria compared to approximately 37\% for both baseline methods. Student assessment data reveal that our concept map-driven approach achieved a significantly lower guess success rate of 28.05\% compared to 37.10\% for the baselines, indicating a more effective assessment of conceptual understanding. The results demonstrate that our concept map-based approach enables robust assessment across cognitive levels and instant identification of conceptual gaps, facilitating faster feedback loops and targeted interventions at scale.
\end{abstract}

\end{frontmatter}

%%%%%%%%%%%%%%%%%%%%%%%%%%%%%%%%%%%%%%%%%%%%%%%%%%%%%%%%%%%%%%%%%%%%%%%%

\section{Introduction}

Multiple choice questions (MCQs), a widely used objective assessment format, require learners to identify the correct answer from options that include plausible but incorrect alternatives, known as distractors \cite{BUTLER2018323}. MCQ-based testing serves two functions: assessment and learning enhancement. Studies have shown that the act of retrieving information while answering MCQs strengthens memory retention and deepens understanding in learners \cite{mcdermott2014both,cantor2015multiple}. Evidence from empirical studies demonstrates that MCQs accelerate learning outcomes. Their objective nature allows straightforward evaluation and rapid feedback, essential for continuous assessments in a lesson plan \cite{thomas2024does,bjork2014multiple}. When distractors are thoughtfully designed to mirror common cognitive errors, MCQs can uncover learning deficiencies and reduce random guessing \cite{gierl2017developing}. This allows educators to modify teaching methods and offer targeted student support \cite{collignon2020alternative}. However, creating quality MCQs with effective distractors that challenge learners at different cognitive levels is resource intensive and requires significant domain expertise \cite{alhazmi2024distractor}. To bridge this gap, we propose a novel framework that utilizes concept maps and instruction-tuned LLMs for automated MCQ generation and assessment. The GitHub repository for the code, concept map, and data is available as a supplementary resource for researchers interested in replicating or building upon our work.\footnote{\href{https://github.com/nicyscaria/AEQG-MCQ-Distractors-Physics}{GitHub Repository}}.

Traditional MCQ generation tackled \textit{question stem} generation, \textit{key answer} identification, and \textit{distractor} generation separately. The generation of \textit{question stems} used rule-based transformations and templates, the identification of \textit{key answers} relied on domain-specific keywords and phrase matching \cite{das2021automatic}, and corpus-based methods focusing on linguistic features or knowledge-based methods for distractors \cite{madri2023comprehensive}. These methods suffered from rigid question templates, a lack of contextual understanding in \textit{key answer} identification, and difficulties in generating plausible semantically relevant distractors, creating major bottlenecks in the creation of high-quality MCQs at scale \cite{alhazmi2024distractor}. Deep learning-based approaches, particularly sequence-to-sequence models \cite{ilya} with attention mechanisms \cite{BahdanauCB14}, offered improvements, particularly the hierarchical recurrent encoder-decoder (HRED) architecture \cite{gao2019generating} with dynamic and static attention \cite{li-etal-2015-hierarchical}.  Subsequent studies explored different attention mechanisms to enhance distractor generation, such as co-attention further \cite{seo2017bidirectional} for passage-question interaction \cite{zhou2020co}, topic-aware attention using Latent Dirichlet Allocation \cite{9533341}, and SoftSel to mitigate answer-revealing distractors \cite{wang2023weak}. 

Large-scale datasets played a crucial role in improving MCQ and distractor generation. While datasets like RACE \cite{zellers-etal-2018-swag} and SWAG \cite{zellers-etal-2018-swag} focused on assessing reading comprehension and reasoning abilities, domain-specific datasets such as SciQ \cite{welbl-etal-2017-crowdsourcing} and EduQG \cite{hadifar2023eduqg} improved structured question generation in different subjects. Distractor selection based on semantic similarity has been done employing cosine similarity of word vectors using these datasets, pre-trained language models (PLMs) such as word2vec \cite{mikolov2013distributed}, GloVe \cite{pennington2014glove}, and fastText \cite{bojanowski2017enriching}. Transformer-based models, such as GPT \cite{radford2019language}, BERT \cite{devlin-etal-2019-bert}, T5 \cite{raffel2020exploring}, and BART \cite{lewis-etal-2020-bart}, further enhanced distractor generation through fine-tuning and retrieval-augmented pre-training approaches. Recent studies leverage in-context learning \cite{mcnichols2023exploring} and single-stage and multistage prompting techniques \cite{doughty2024comparative,maity2024novel} to improve distractor quality. These investigations have identified significant room for improvement in the generation of domain-specific and cognitively diverse questions with effective distractors. Furthermore, while promising, LLMs introduce new challenges such as bias \cite{gallegos-etal-2024-bias} and hallucination \cite{zhang2023siren}, which could lead to seemingly accurate, yet factually inaccurate \textit{key answers} and \textit{distractors}. Although knowledge integration shows promise in grounding model output, current methods still struggle to fully address these challenges, necessitating expert validation, which creates significant resource bottlenecks.

Addressing these challenges, particularly the need for effective distractors and reduced expert intervention, we propose a comprehensive framework for MCQ generation and evaluation that integrates hierarchical concept maps with an instruction-tuned LLM for the generation of MCQs with appropriate distractors. The framework also automatically validates the correctness of the generated question, answer, and distractors. Our key contributions are threefold. First, we developed a hierarchical concept map of classical physics covering high school content that organizes domain knowledge across grade, unit, topic, and subtopic levels to enable systematic knowledge retrieval and integration. Second, we developed a framework that leverages this concept map structure to guide LLM-based MCQ generation, implementing automated context retrieval while ensuring question diversity and preventing repetition with a validation step to ensure the correctness of the question, answer, and distractors. Third, we provide empirical validation through student assessments and expert evaluations to compare our framework with baseline LLM and RAG-based approaches, focusing on how concept map integration improves MCQ quality and reduces the need for expert intervention through automated evaluation.

In the following sections, we detail our automated MCQ generation methodology (Section~\ref{methodology}), followed by expert evaluation (Section~\ref{expert_eval}) and learner-centric evaluation (Section~\ref{learner_eval}), each covering their respective methodologies and analysis. We then discuss key findings, practical implementation considerations, and future research directions (Section~\ref{discussion}), and conclude with broader implications of our work (Section~\ref{conclusion}).

%%%%%%%%%%%%%%%%%%%%%%%%%%%%%%%%%%%%%%%%%%%%%%%%%%%%%%%%%%%%%%%%%%%%%%%%

\section{Automated MCQ Generation Methodology}\label{methodology}

\begin{figure*}[h]%
\centering
\includegraphics[width=0.8\textwidth]{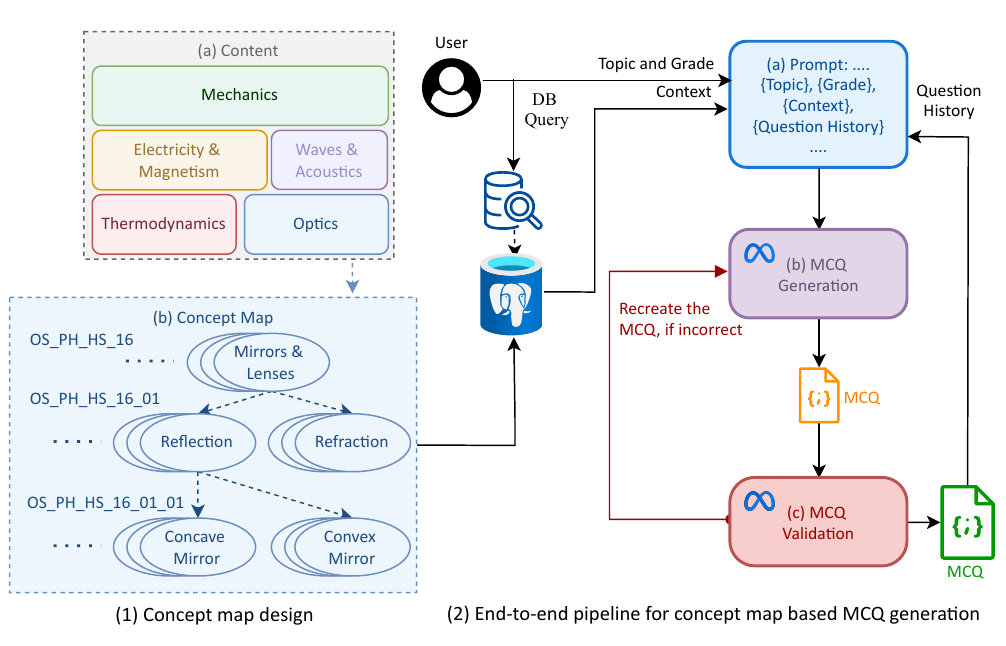}
\caption{Schematic of the hierarchical concept map-based MCQ generation and its validation.}\label{block_diagram}
\vspace{0.5cm}
\end{figure*}

Figure~\ref{block_diagram} shows the schematic of the overall system we developed for automated generation and validation of MCQs. The key components are the concept map, LLM-based question generation from the concept map, and automated validation.

\subsection{Content and Concept Map Design}

The concept map is the central piece of our automated MCQ generation framework. To systematically organize the content, we created a hierarchical concept map encompassing 19 units of classical physics using the OpenStax High School Physics textbook \cite{urone_hinrichs_2020}. These units encompassed five major topics in physics (Figure~\ref{block_diagram}(1a)): Mechanics, Electricity and Magnetism, Waves and Acoustics, Thermodynamics, and Optics. This hierarchical framework represents a structured knowledge repository to accommodate future expansion with additional subjects and grade levels.

Within our concept map (Figure~\ref{block_diagram}(1b)), each `unit' is aligned with a unit from the OpenStax textbook, encapsulating `topics' that are further divided into `subtopics'. `Topics' align with sections in the OpenStax textbook, and for each topic, we mapped learning objectives to the cognitive and knowledge dimensions of the revised Bloom's taxonomy \cite{blooms}. This mapping ensures that the knowledge structure not only captures content but also the cognitive processes appropriate for instruction and assessment. We consider question difficulty through cognitive levels of Bloom's taxonomy, where higher-order thinking skills are considered more difficult than lower-order skills. At the most granular level of the hierarchy, subtopics have seven attributes: \textit{(i)} prerequisites, \textit{(ii)} mathematical formulations, \textit{(iii)} common misconceptions, \textit{(iv)} engineering applications, \textit{(v)} cross-cutting topics, \textit{(vi)} analogies, and \textit{(vii)} alignment with India's National Council of Educational Research and Training (NCERT) curriculum (presented as linked references to pertinent sections in NCERT textbooks for grades 8-12). 

The prerequisites are divided into physics-specific concepts (e.g., understanding velocity before acceleration) and external knowledge from other subjects (such as mathematical skills), allowing the system to account for knowledge dependencies. The subtopics contain detailed mathematical formulations corresponding to the concept. For example, an equation such as the one for kinetic energy is represented as a JSON object that includes the equation itself formatted using LaTeX syntax (e.g., \verb|$KE = \frac{1}{2} m v^2$|), which allows for accurate mathematical rendering upon retrieval. This data object also contains definitions for each variable and the SI units involved in the equation. Common misconceptions identified through literature and teaching experience inform the creation of targeted assessment questions. Cross-cutting topics highlight connections between physics concepts and other disciplines, enabling interdisciplinary assessment. Engineering applications provide a real-world context that demonstrates the practical relevance of theoretical concepts. Carefully selected analogies help bridge abstract physics concepts with familiar experiences. 

The content was reviewed by physics subject matter experts and iteratively refined to ensure accuracy and pedagogical appropriateness. Building the physics concept map required approximately 80 expert-hours (with 4 subject-matter experts each spending 20 hours), including creation and validation of the concept map. We experimented with creating the concept map in a semi-automated way using larger and expensive LLM APIs with in-context examples and found that this could reduce expert effort, though expert validation remains essential. While our current work focuses on physics, the hierarchical framework with grades, subjects, units, topics, and subtopics with detailed information is domain-agnostic. In future iterations, we plan to align with the Next Generation Science Standards (NGSS) \cite{NGSS}. This comprehensive hierarchical organization of concept maps provides the foundation for our automated MCQ generation system to create contextually and pedagogically aligned MCQs. Beyond question generation, this concept map represents a valuable one-time investment with multiple educational applications. The structured knowledge enables intelligent tutoring systems by mapping learning pathways based on prerequisites, supports targeted remediation using documented misconceptions, informs curriculum design, and enhances educational analytics. Structured knowledge also facilitates cross-disciplinary connections by explicitly linking physics concepts to other STEM subjects.

The framework's database architecture is another key component of our system, specifically engineered for efficient knowledge management and retrieval. A relational SQL database (PostgreSQL) provides the structure and maintains hierarchical relationships between knowledge elements. Significant flexibility is introduced via a hybrid storage model: standard VARCHAR fields store basic identifiers (like keys and names), whereas complex pedagogical information uses JSON. This targeted use of JSON enables the database to easily manage the varied and evolving content types within the overall relational framework. This structured database approach offers considerable efficiency advantages compared to common RAG methods, which typically perform semantic searches across large text datasets. Direct querying of the database using SQL, enhanced by indexing JSON fields, allows for faster and less computationally intensive retrieval of specific, targeted information compared to semantic searching over document embeddings. Additionally, this structured storage method generally requires less storage capacity, as it holds distilled knowledge elements rather than the voluminous and potentially duplicated text segments often necessary for RAG systems. This database design, therefore, supports robust information retrieval while improving speed and reducing the computational load during the question generation phase.

\subsection{Selection of Generative Language Model}

Our concept-map-based MCQ generation framework requires a capable LLM for the generation task. We conducted a comprehensive evaluation of different open-source and proprietary generative language models,  focusing on their MCQ generation performance while considering the practicality of API costs for potential production deployment. We evaluated the following six models: Llama 3.1 70B, Llama 3.3 70B model \cite{dubey2024llama}, Qwen 2 72B \cite{bai2023qwen}, GPT-4o \cite{hurst2024gpt}, GPT-4o mini, and Gemini 1.5 Pro \cite{team2024gemini}. 

To systematically assess model performance, we designed a controlled experiment using five diverse mechanics topics. For each topic, we generated questions across the cognitive spectrum from the `Remember' level to the `Evaluate' level of Bloom's taxonomy, resulting in 25 questions per model. Each generated question was manually evaluated for scientific accuracy and factual correctness, focusing specifically on whether the physics concepts were correctly represented and whether exactly one answer option was correct. Our analysis revealed that Llama 3.3 70B outperformed other models, producing fewer conceptual errors compared to the alternatives we tested. This performance advantage, combined with its favorable cost profile for deployment scenarios, made Llama 3.3 70B the optimal choice for our framework. Although proprietary models like GPT-4o occasionally produced more elegantly phrased questions, the accuracy advantage of Llama, 3.3 70B is more critical for educational applications. LangChain was used as the development framework to facilitate experimentation with prompt templates and maintain consistency.

\subsection{MCQ Generation Methods}

After experimenting with various prompting strategies for each approach, we developed a structured Chain-of-Thought (CoT) prompt \cite{wei2022chain} that specifies all generation requirements. The prompt template incorporates four key inputs: the Physics topic, grade level, targeted cognitive skill level from Bloom's taxonomy, and definition of the skill. The prompt also enforced key pedagogical requirements: (1) each distractor should address a misconception or a prerequisite specific to the topic, (2) the student should be able to answer the question only if they have mastered the concept, (3) language complexity must align with the student's grade level, and (4) questions must include context-specific physics applications. The prompt specifies an output JSON format to facilitate structured data handling, capturing the question text, the target skill, four answer options, the correct answer, detailed explanations of the correct answer, and the specific misconception or prerequisite tested by each distractor.

\begin{algorithm*}
\caption{Concept Map-Based MCQ Generation Pipeline}
\label{alg:mcq_generation}
\begin{algorithmic}[1]
\Require User input $u$, Grade level $g$, Cognitive skills $\mathcal{S} = \{\text{Remember, Understand, Apply, Analyze, Evaluate}\}$
\Ensure Assessment with validated MCQs $\mathcal{Q} = \{q_1, q_2, ..., q_{|\mathcal{S}|}\}$ in JSON format

\State $t_{user} \leftarrow \textsc{ExtractTopic}(u)$ \Comment{Extract physics topic from user input using LLM}
\State $\mathcal{T} \leftarrow \textsc{RetrieveTopics}(g)$ \Comment{Get all topics for grade level from concept map}
\State $t_{matched} \leftarrow \textsc{LLMMatch}(t_{user}, \mathcal{T}, \text{temperature}=0)$ \Comment{Zero-shot topic matching}
\State $query \leftarrow \textsc{GenerateSQLQuery}(t_{matched})$ \Comment{Create database query}
\State $context \leftarrow \textsc{ExecuteQuery}(query)$ \Comment{Retrieve structured knowledge}
\State $\mathcal{Q} \leftarrow \{\}$ \Comment{Initialize assessment}

\For{each $s \in \mathcal{S}$} \Comment{Generate questions for each cognitive skill level}
    \State $q \leftarrow \textsc{GenerateValidQuestion}(s, t_{user}, context, \mathcal{Q})$ \Comment{Generate valid question for skill level}
    \If{$q \neq \text{NULL}$}
        \State $\mathcal{Q} \leftarrow \mathcal{Q} \cup \{q\}$ \Comment{Add valid question to assessment}
    \EndIf
\EndFor

\State \textbf{return} $\mathcal{Q}$

\vspace{0.3cm}

\Function{GenerateValidQuestion}{$s, t, context, \mathcal{H}$}
    \State $max\_attempts \leftarrow 2$ \Comment{Prevent infinite loops}
    \For{$attempt = 1$ to $max\_attempts$}
        \State $question\_history \leftarrow \{q.\text{question} : q \in \mathcal{H}\}$ \Comment{Extract previous questions}
        \State $prompt \leftarrow \textsc{BuildPrompt}(context, s, question\_history)$ \Comment{Construct generation prompt}
        \State $q \leftarrow \textsc{LLMGenerate}(prompt, \text{temperature}=0.75, \text{CoT}=\textbf{true})$ \Comment{Generate MCQ with Chain-of-Thought}
        
        \If{\textsc{ValidateJSONFormat}($q$)} \Comment{Check JSON format compliance}
            \State $eval \leftarrow \textsc{EvaluateQuestion}(q, question\_history)$ \Comment{Detailed evaluation}
            
            \If{$eval.\text{valid}$}
                \State \textbf{return} $q$ \Comment{Question passes all checks}
            \Else
                \State $q_{fixed} \leftarrow \textsc{FixQuestion}(q, eval, question\_history)$ \Comment{Attempt to fix issues}
                \State $eval_{fixed} \leftarrow \textsc{EvaluateQuestion}(q_{fixed}, question\_history)$
                \If{$eval_{fixed}.\text{valid}$}
                    \State \textbf{return} $q_{fixed}$
                \EndIf
            \EndIf
        \EndIf
    \EndFor
    \State \textbf{return} NULL \Comment{Failed to generate valid question}
\EndFunction

\vspace{0.2cm}

\Function{EvaluateQuestion}{$q, \mathcal{H}$}
    \State Evaluate uniqueness: $unique \leftarrow \textsc{CheckUniqueness}(q, \mathcal{H})$
    \State Evaluate correctness: $correct \leftarrow \textsc{CheckAnswerCorrectness}(q)$
    \State $valid \leftarrow unique \land correct$
    \State \textbf{return} $\{valid, unique, correct, issues\}$
\EndFunction

\vspace{0.2cm}

\Function{FixQuestion}{$q, eval, \mathcal{H}$}
    \If{$\neg eval.unique$}
        \State \textbf{return} \textsc{GenerateNewQuestion}($q.skill$, $\mathcal{H}$) \Comment{Generate completely new question}
    \ElsIf{$\neg eval.correct$}
        \State \textbf{return} \textsc{FixAnswerOnly}($q$) \Comment{Fix only answer and explanation}
    \EndIf
    \State \textbf{return} $q$
\EndFunction

\end{algorithmic}
\end{algorithm*}

\subsubsection{Baseline Methods}
We implemented two baseline methods to benchmark our new concept map-based MCQ generation.

In the first baseline, henceforth called the ``base LLM approach", we used the Llama 3.3 70B Instruct model accessed through the TogetherAI API services\footnote{\url{https://api.together.ai/}}.  A temperature setting of 0.75 balances creativity in question generation and maintains consistency and coherence in the output.

In the second baseline, henceforth called the ``RAG-based approach", we constructed a vector database using ChromaDB.\footnote{\url{https://www.trychroma.com/}} To populate this database, we generated embeddings of the entire content of the OpenStax High School Physics textbook (excluding exercise sections) using the all-mpnet-base-v2 \cite{song2020mpnet} model from sentence transformers \cite{sentence}. During question generation, the system retrieves the three most similar content chunks using similarity search, which are then provided as a context to the LLM. The RAG-based prompt extended the base prompt by incorporating this retrieved textbook content.

\subsubsection{Concept Map-Based MCQ Pipeline}\label{concept-map_desc}

Figure~\ref{block_diagram}(2) illustrates our end-to-end pipeline for concept map-based MCQ generation, consisting of three main components: (a) prompting with structured content, (b) MCQ generation, and (c) validation. The pipeline is also detailed as an algorithm in Algorithm~\ref{alg:mcq_generation}.

The process starts by finding the topic in our database that is most similar to the user's input. This matching uses an LLM in a zero-shot manner, i.e., without specific examples and with a temperature set to zero for deterministic results.  Specifically, the LLM is presented with a list of all the topics within the specific grade level in our concept map and asked to find the closest match to the user's topic.  After identifying a matching topic, the system generates and executes precise SQL queries to retrieve related subtopics. The data retrieved includes the description of the subtopics, mathematical formulations, prerequisites, misconceptions, cross-cutting topics, engineering applications, and analogies. This structured information then becomes the context for the LLM, which it uses to generate MCQs that are relevant to the user's topic and aligned with the curriculum.

This structured context feeds into the MCQ generation component (Figure~\ref{block_diagram}(2b)), where the LLM creates questions for each cognitive skill level, progressing from `Remember' to `Evaluate' based on the provided context and Chain-of-Thought prompt. Our concept map-based prompt (Figure~\ref{block_diagram}(2a)) was the most sophisticated, explicitly instructing the model to use specific structured knowledge elements extracted as the context and to avoid generating questions similar to the question history. 

The generated MCQs then undergo validation (Figure~\ref{block_diagram}(2c)) through two steps: first, verify adherence to the specified JSON format, and second, an automated validation in which the LLM acts as a judge \cite{zheng2024judging} to check the correctness of the question and its answer. If either validation step fails, the system repeats the generation process with the same input until a valid question is produced. A key challenge we addressed was preventing repetitive question scenarios at different cognitive levels. Since questions are generated sequentially for each skill level, there was a tendency for the LLM to reuse similar contextual scenarios within the same topic. To mitigate this issue, we implemented a question history tracking mechanism. Successfully validated questions are added to this history, serving as an additional context during subsequent question generation. This approach explicitly instructs the LLM to create new scenarios that differ from previously generated questions, ensuring diversity in the practical applications, scenarios, and contexts used across the cognitive spectrum. This diversity is important to maintain student engagement and comprehensively assess different aspects of conceptual understanding.

To evaluate our framework and create a benchmark dataset, we generated questions across 50 topics from the OpenStax textbook using all three approaches. For each topic, we generated five questions that target five levels of cognitive ability, resulting in a dataset of 750 MCQs. This dataset, which we call the \textit{\textbf{OpenStax PhyQ}} dataset, follows our specified JSON format and is publicly available to support future research in the GitHub Repository\footnote{\href{https://github.com/nicyscaria/AEQG-MCQ-Distractors-Physics}{GitHub Repository}}.

\section{Expert Evaluation}\label{expert_eval}

We performed an expert evaluation to assess the quality, relevance, and pedagogical soundness of the generated MCQs.

\subsection{Method}
A panel of four experts with graduate-level education and subject matter expertise in physics conducted the evaluations. Each question was independently evaluated by two of our four subject matter experts using an eight-item rubric (Table~\ref{table:evaluation-criteria}) that assessed both question quality and distractor effectiveness. The experts followed a hierarchical evaluation approach. If the response to `Relevance' or `Correctness' was `No', the evaluation of the question was discontinued, and the remaining criteria were marked as `NA', as incorrect or irrelevant questions would be unsuitable for assessment purposes. This approach optimized the evaluation process by focusing the effort of experts on the questions that met these fundamental requirements.

\begin{table*}[h]
\caption{Hierarchical eight-item rubric used to evaluate generated MCQs}
\label{table:evaluation-criteria}
\centering
\vspace{0.5cm}
\begin{tabular}{ll}
\toprule                 
Rubric item & Definition and response option\\
\midrule
\multicolumn{2}{c}{\textbf{Question Evaluation}} \\
\midrule
Relevance & Is the question relevant to the specified Physics topic given? \textit{(yes/no)}\\
Correctness  & Are the question and its designated answer scientifically accurate? \textit{(yes/no)}\\
GradeLevel & Is the complexity, language, and content suitable for the target grade level?  \textit{(yes/no)}\\
Similarity  & Is the question unique compared to questions generated on the same topic? \textit{(yes/no)}\\
BloomsLevel & What is the Bloom’s skill associated with the question? \textit{(remember, understand, apply, analyze, evaluate, and create)} \\
\midrule
\multicolumn{2}{c}{\textbf{Distractor Evaluation}} \\
\midrule
Plausibility  & Is the distractor believable enough to confuse a partially informed student? \textit{(yes/no)}\\
Misconceptions  & Does the distractor address common misconceptions or missing prerequisites? \textit{(yes/no)}\\
Independence & Are the distractors truly distinct from each other, avoiding overlap or logical dependencies? \textit{(yes/no)}\\
\bottomrule
\end{tabular}
\end{table*} 

\begin{table*}[h]
    \caption{Expert evaluation: Inter-annotator agreement on the hierarchical rubric for generated MCQs across all methods of generation.}
    \label{tab:agreement_kappa}
    \centering    
    \vspace{0.5cm}
    \begin{tabular}{lcccccccc}
     \toprule
        & Relevance & Correctness & GradeLevel & Similarity & Bloom'sLevel & Plausibility & Misconceptions & Independence \\
        \midrule
        Agreement & 1.00 & 1.00 & 0.99 & 0.96 & 0.90 & 0.96 & 0.98 & 0.99 \\
        Kappa & 1.00 & 1.00 & 0.83 & 0.86 & 0.86 & 0.62 & 0.56 & 0.58 \\
        \bottomrule
    \end{tabular}
\end{table*}

Our rubric intentionally addressed specific quality dimensions. The `Relevance criterion ensured that the questions targeted the specified physics topic. `Correctness' verified the scientific accuracy of both the question formulation and the designated correct answer. `GradeLevel' assessment examined whether vocabulary, sentence structure, and concept complexity matched the educational stage of the target audience. `Similarity' checks prevented redundancy, ensuring that questions on the same topic approached the concept from different angles. The `BloomsLevel' rating identified which cognitive skill was being tested, allowing us to evaluate if the questions adequately addressed their intended taxonomic level.

For distractor evaluation, we focused on three criteria. `Plausibility' ensured that distractors were not obviously wrong but would challenge students with incomplete understanding. `Misconceptions' assessment verified that each distractor targeted specific conceptual errors or knowledge gaps common in physics education. `Independence' verified that the distractors represented distinct misunderstandings rather than overlapping or logically dependent errors.

For each evaluation criterion, other than `BloomsLevel' alignment, a question was considered acceptable only when both expert reviewers responded with a `Yes'. For the distractor-focused criteria, both experts were required to respond `Yes' for all distractors within a given question for that question to be considered acceptable on those criteria. A generated MCQ was ultimately classified as `high quality' only when it received unanimous `Yes' responses from both expert reviewers in all evaluation criteria. For questions where the experts' initial `BloomsLevel' alignment differed, the two raters engaged in a detailed discussion to arrive at a final consensus on categorizing the assigned cognitive skill level.

Human judgment naturally varies, even among subject matter experts examining the same content. These differences in expert evaluations can stem from personal factors such as writing style, assumptions, knowledge, and attention to detail \cite{amidei2018rethinking}. To maintain evaluation consistency and address rating variances, we assessed inter-rater reliability with various metrics. We determined the percentage agreement for all criteria, offering a simple view of identical judgments by raters. However, since it overlooks the possibility of chance agreement, we also calculated Cohen's Kappa ($\kappa$) \cite{mchugh2012interrater} for binary criteria to give a more robust agreement measure. For the ordinal Bloom's taxonomy evaluations, we applied quadratic weighted Cohen's Kappa \cite{cohen1968weighted}, which appropriately penalizes larger disagreements (like between `Remember' and `Analyze') more heavily than disagreements between adjacent cognitive levels. This systematic approach of measuring agreement ensured our assessment process remained valid despite the natural variations in expert reviews.

\subsection{Analysis}

Table \ref{tab:agreement_kappa} presents the inter-annotator agreement for each criterion between the two expert reviewers. The expert judgements were highly reliable, with perfect agreement (1.00) for Relevance and Correctness and high agreement (0.90-0.99) for other criteria. Cohen's Kappa values indicated strong agreement ($\kappa > 0.80$) for the question-related criteria (Relevance, Correctness, GradeLevel, Similarity, and Bloom's Level) and moderate to fair agreement for distractor-related criteria (Plausibility, Misconception and Independence), likely due to the subjectivity in evaluating plausibility and identifying misconceptions.

Table \ref{tab:performance_metrics} shows the performance of the different MCQ generation methods (LLM, RAG, and concept map). The `High Quality' row indicates the percentage of questions that received unanimous positive evaluations from both experts across all assessment criteria. The results are expressed as the percentage of questions that met the specified criteria, based on agreement between the two experts.

\begin{table}[h]
\caption{Expert evaluation: Performance of the different methods of MCQ generation on the hierarchical rubric for questions and distractors.}
    \centering
    \vspace{0.5cm}
    \label{tab:performance_metrics}
    \begin{tabular}{lccc}
        \toprule
        & LLM & RAG & Concept Map \\
        \midrule
        Relevance & \textbf{98.80\%} & 94.80\% & 97.20\% \\
        Correctness & 75.60\% & 78.00\% & \textbf{88.00\%} \\
        GradeLevel & 74.80\% & 77.20\% & \textbf{86.80\%} \\
        Similarity & 54.80\% & 50.00\% & \textbf{87.60\%} \\
        Bloom'sLevel & 73.20\% & 73.60\% & \textbf{78.80\%} \\
        \midrule
        Plausibility & 60.80\% & 66.00\% & \textbf{80.00\%} \\
        Misconceptions & 64.00\% & 66.80\% & \textbf{82.80\%} \\
        Independence & 67.60\% & 68.40\% & \textbf{83.60\%} \\
        \midrule
        High Quality & 37.60\% & 37.20\% & \textbf{75.20\%} \\
        \bottomrule
    \end{tabular}
\end{table}

In question-focused criteria, all MCQ generation methods demonstrated strong performance in `Relevance', with scores consistently exceeding 94\%. However, `Correctness' showed more notable variation, with our concept map-based generation achieving the highest score around 88\%, while the baseline RAG and LLM approaches performed moderately lower in the mid-70s range. `Grade level' appropriateness followed a similar pattern, with concept map-based generation leading at nearly 87\%, followed by RAG and LLM approaches at around 75\%. The most pronounced difference emerged in `Similarity' assessment, where concept map-based generation significantly outperformed (around 88\%) both LLM and RAG based approaches (around 50\%), highlighting a critical weakness in generating diverse questions for these baseline methods. `BloomsLevel' alignment remained relatively consistent across methods, with our concept map-based generation maintaining a slight lead at around 79\%.

In distractor-focused criteria, concept map-based generation consistently demonstrated superior performance across all metrics: achieving 80\% in `Plausibility', nearly 83\% in addressing common `Misconceptions', and approximately 84\% in maintaining `Independence' between options. In contrast, the baseline RAG and LLM approaches performed notably lower, with the RAG-based generation scoring slightly better (66-68\% in all metrics) than the base LLM approach (61-68\% range). Specifically, the base LLM approach showed its weakest performance in `Plausibility' at around 61\%, while performing marginally better in `Independence' at 68\%. RAG approach maintained a more consistent performance across these criteria, with scores hovering around 66-68\%, but still significantly behind the concept map-based generation. These results highlight the particular strength of our concept map-based method in generating pedagogically sound distractors that are not only plausible enough to challenge students but also effectively address common misconceptions while maintaining clear distinctions between options.

The aggregate performance metric, represented by the `High Quality' row of Table~\ref{tab:performance_metrics}, provides a comprehensive measure of MCQ quality by calculating the percentage of questions that successfully met all evaluation criteria. Our new concept map-based MCQ generation and validation approach demonstrated remarkable superiority, achieving a 75.20\% success rate, indicating that three out of four questions satisfied every quality criterion. In stark contrast, both the base LLM and RAG approaches achieved notably lower success rates of approximately 37\%, indicating that barely more than one-third of their generated questions met all quality standards. The results demonstrate the clear superiority of the concept map-based approach over baselines in generating high-quality MCQs. The evaluation process's reliability, supported by strong inter-annotator agreement, validates these findings and demonstrates that incorporating structured domain knowledge through concept maps significantly enhances MCQ generation quality.

\section{Learner Centric Evaluation}\label{learner_eval}

To complement our expert evaluation, we conducted a learner-centric evaluation with students to assess how effectively each generation method created questions that measured true understanding while resisting successful random guessing.

\subsection{Method}
An evaluation was performed with 145 ninth-grade students from an Indian public school to assess the quality and effectiveness of the automatically generated MCQs. This school accommodates students from diverse socioeconomic backgrounds and regions, serving primarily children of central government employees who are often transferred nationwide.

The questions for the learner-centered evaluation were randomly selected from the generated MCQs marked as `Yes' by both expert evaluators on the `Correctness' rubric. This initial review verified that each question was scientifically accurate and had precisely one correct answer among the options provided. No additional quality checks were performed besides this basic correctness verification, as our goal was to assess the unfiltered output quality of each generation method.

Each student completed a 15-question test designed to include five questions from each of the three generation methods. We selected questions from various Grade 9 Physics topics for the tests. Within each test, questions were arranged to progressively increase cognitive skill level, from `Remember' to `Evaluate' based on Bloom's taxonomy. The test design carefully ensured that no unique combination of topic, skill level, and generation method appeared more than once within any test version. This prevented repetition and ensured that students encountered diverse questions in their assessment. Across the student population, nine unique test versions were administered, each student taking only one version. This approach allowed us to cover a broader range of generated questions while keeping the test length manageable for individual students. For each question answered, we collected the student response, response time, self-reported guessing, and perceived difficulty. Students directly indicated whether they had guessed the answer by selecting `Yes' or `No' for self-reported guessing. They also rated each question's perceived difficulty as `Difficult', `Moderately Difficult', or `Easy'.

The main objective was to analyze the learners' guessing behavior across three MCQ generation techniques, aiming to assess how well the distractors generated deterred successful random guesses. `Guessing Success Rate' was the main metric used, with `Accuracy' and `Difficulty Weighted Accuracy' used as additional metrics to provide a comprehensive evaluation of question quality. The metrics are defined as follows. 
\begin{equation}
\text{Accuracy} = \frac{\sum_{i=1}^{N} C_i}{\sum_{i=1}^{N} A_i} \times 100
\end{equation} 

\begin{equation}
    \text{Guessing Success Rate} = \frac{\sum_{i=1}^{N} (G_i \times C_i)}{\sum_{i=1}^{N} G_i} \times 100
\end{equation} 

\begin{equation}
\text{Difficulty Weighted Accuracy} = \frac{\sum_{i=1}^{N} (C_i \times D_i)}{\sum_{i=1}^{N} D_i} \times 100,  
\end{equation}
where $C_i = 1$ if question $i$ was answered correctly, 0 if incorrect, $A_i = 1$ if question $i$ was attempted, $G_i = 1$ if question $i$ was self-reported as guessed, 0 if not guessed and $D_i$ represents the difficulty level of question $i$ as perceived by students, with values of 5 for `Difficult', 3 for `Moderately difficult', and 1 for `Easy'. 

This comprehensive evaluation methodology, which combined expert assessment through structured criteria and learner-centered evaluation through student performance metrics, provided a rigorous assessment of MCQs generated by our concept map-based approach and the two baseline approaches.

\subsection{Analysis}

The learner-centric evaluation revealed notable differences in performance in the three question generation methods, as shown in Table~\ref{table:accuracy_metric}. In terms of accuracy, the base LLM approach exhibited the highest performance at 44.47\% with the RAG-based and concept map-based approaches following at 40.66\% and 37.25\% respectively. This pattern was consistent when accounting for question difficulty in our Difficulty Weighted Accuracy metric, where the base LLM approach again led with 41.08\%, followed by RAG (37.67\%) and concept map approaches (33.60\%).  

The concept map-based generation demonstrated particularly interesting characteristics with respect to the difficulty of the question and the student's guessing behavior. Although it showed the lowest raw accuracy (37.25\%) and difficulty-weighted accuracy (33.60\%), this appears to be directly related to the challenging nature of its questions. The concept map-based questions leveraged specific misconceptions and prerequisite knowledge from our structured knowledge repository, creating more nuanced challenges for students that required deeper conceptual understanding.

This interpretation is strongly supported by the analysis of Guessing Success Rate, where the concept map-based generation showed the lowest rate (28.05\%), compared to RAG (33.16\%) and the base LLM approach (37.10\%). This significantly lower Guessing Success Rate indicates that the questions generated using the concept map-based approach were more resistant to successful guessing strategies. When students were uncertain and resorted to guessing, they were less likely to choose the correct answer for concept map-generated questions than for questions created by the other two methods.

\begin{table}[h]
\caption{Performance metrics across different generation methods (lower the better)}
\vspace{0.65cm}
\label{table:accuracy_metric}
\centering
\begin{tabular}{l@{\hspace{2em}}c@{\hspace{2em}}c@{\hspace{2em}}c}
\toprule
\textbf{Metric (\%)} & \textbf{LLM} & \textbf{RAG} & \textbf{Concept Map} \\
\midrule
Accuracy & 44.47 & 40.66 & \textbf{37.25} \\
Difficulty Weighted Accuracy & 41.08 & 37.67 & \textbf{33.60} \\
Guessing Success Rate & 37.10 & 33.16 & \textbf{28.05} \\
\bottomrule
\end{tabular}
\end{table}

To rigorously analyze the effectiveness of the distractors and confirm this initial observation, we conducted a chi-square test of homogeneity. This test revealed a significant relationship between the question generation method and guessing success ($\chi^2 = 6.78$, $p = 0.034$), indicating that the question generation method influenced how successfully students could guess correct answers. To pinpoint specific differences, we then performed pairwise comparisons using z-tests. These tests highlighted that the concept map-based generation exhibited a guessing success rate of 28.05\%, which was significantly lower than the base LLM approach's 37.10\% ($p = 0.009$) at $\alpha = 0.05$.  Crucially, this significant difference persisted even after applying the more conservative Bonferroni correction ($\alpha= 0.0167$), strengthening the robustness of this finding. The RAG-based approach, with an intermediate guess success rate of 33.16\%, did not show statistically significant differences compared to the LLM ($p = 0.235$) or concept map approaches ($p = 0.145$), both before and after Bonferroni correction. We also analyzed response times across all three methods and found them to be comparable, indicating that the differences in performance were not attributable to students spending more or less time on questions from any particular method.

These findings suggest that the concept map-based generation, while producing questions with lower accuracy scores, may be generating more discriminating assessment items that better differentiate between students who genuinely understand the physics concepts and those who do not. The structured knowledge in the concept map appears to enable the creation of more challenging distractors that effectively target common misconceptions, making it harder for students to guess correctly without true conceptual mastery. This result is significant because it shows that while the learners scored lower on the concept map-based questions, these questions were more challenging, making them difficult to guess correctly. 

Although these results are promising and point to distinct advantages of the concept map-based approach in distractor generation, a larger sample size would provide greater statistical power to detect potential differences between the methods. Furthermore, expanding the study to include students at multiple grade levels would help establish the generalizability of these findings. 

\section{Discussion}\label{discussion}

Our evaluation revealed that structured domain knowledge significantly improves MCQ quality, particularly in terms of pedagogical and distractor effectiveness. Our method significantly outperformed the baseline methods, with three out of four questions meeting all quality criteria, whereas the baseline methods achieved just over one third. The concept map-based generation excelled in expert evaluation metrics compared to LLM and RAG approaches. Statistically significant differences were found in learner-centric evaluations for guessing success rates. Our concept map-based MCQs had lower guessing success compared to the base LLM, suggesting that they were more challenging, less prone to random guessing, and better at assessing deep conceptual understanding rather than surface knowledge. These results underscore the fundamental advantages of structured knowledge representation in educational content generation. While stronger LLMs might improve baselines, structured grounding through concept maps provides distinct advantages that extend beyond model capabilities. Concept map creation represents a one-time investment that enables consistent, pedagogically sound generation—a process that could be semi-automated with a capable LLM and some in-context examples. Without such a structured foundation, even expensive models require extensive prompting and often produce distractors that could increase guessing success rates, undermining the assessment's effectiveness.

Our analysis yielded encouraging results and intriguing insights. Base LLMs showed correct reasoning, yet occasionally chose incorrect answers, whereas RAG systems at times retrieved tangential or irrelevant data. The concept map-based generation approach guaranteed relevant content due to its deterministic nature. Notably, while all three methods performed consistently across various physics topics, the concept map-based approach demonstrated better consistency at higher cognitive levels compared to the other methods. Nevertheless, all approaches still exhibited some decline in performance as cognitive complexity increased, particularly at the highest levels of Bloom's taxonomy.

\subsection{Practical Implementation Considerations}

Beyond effectiveness, our approach offers significant practical advantages in deployment scenarios. The concept map-based system demonstrates notable cost savings compared to the alternatives. Unlike RAG-based approaches that require substantial storage for text embeddings and computationally expensive similarity searches, our SQL database implementation is lightweight and deterministic. API-based deployment of the system is straightforward, as it does not require model fine-tuning or specialized hardware. This makes our approach particularly suitable for resource-constrained educational environments. Additionally, high-school physics content remains relatively stable over time, making the initial concept map creation a one-time investment with long-term returns.

\subsection{Limitations and Future Work}

All approaches faced challenges with complex mathematical tasks, such as three-digit multiplication or division and inverse trigonometric functions, as well as in balancing theoretical with numerical content and aligning questions with the targeted cognitive levels. Difficulties were encountered with strictly numerical queries, often resulting in questions limited to the application level even when higher cognitive levels were intended. These findings can be attributed to the inherent limitations of LLMs in performing mathematical tasks.

Future studies might extend the evaluation to a broader student group in various grades to enhance the applicability of our results. Further technical improvements could focus on advancing mathematical processing, possibly via specialized modules, and refining validation, especially for complex numerical computations.

\section{Conclusion}\label{conclusion}

A new concept map-based method with LLMs was developed for generating high-quality MCQs with distractors that reduce successful guessing in high school physics education. Our approach outperformed the baseline LLM and RAG systems in expert evaluations of the generated MCQs, with three-quarters of the concept map-generated questions meeting all quality criteria compared to only one-third of the baseline methods. Student testing further validated these findings, revealing lower guessing success rates for our method. 

Our work's implications extend beyond MCQ generation, providing a comprehensive physics concept map and an expert-validated MCQ dataset with assigned Bloom's taxonomy levels. Our structured knowledge representation methodology could serve as a template for expanding to other subject areas using more capable models like OpenAI O1 or Claude 3.5 Sonnet, significantly reducing the effort required to develop similar systems for other subjects. 

Furthermore, our question format deliberately links misconceptions to distractor options, creating a powerful diagnostic tool for educators. When a student selects a specific incorrect answer, this choice is directly mapped to a particular misconception or knowledge gap identified in our concept map. This structured approach transforms multiple-choice assessments from simple evaluation tools to rich diagnostic instruments. For example, if a student consistently selects distractors related to confusing velocity with acceleration in different questions, educators can precisely identify this conceptual misunderstanding and provide targeted remediation. This diagnostic capability extends beyond individual students to classroom-level analysis, allowing teachers to identify common misconceptions between groups and adjust instruction accordingly. By connecting incorrect responses to specific conceptual gaps, our approach enables more efficient and effective educational interventions tailored to students' actual learning gaps rather than generic remediation.

Our work shows the promise of structured knowledge representation in MCQ creation, but further progress in cognitive alignment and validation is essential to unlock the full potential of automated STEM assessments. Intelligent tutoring systems can use our automated MCQ generation to develop customized assessments for students at scale, with the potential to revolutionize access to high-quality education across the world, especially in developing economies.

\section*{Ethics Statement}

This study involved student participants under the age of 18. Prior to data collection, informed consent was obtained from parents or legal guardians by the school teacher. The assessment was administered during regular school hours under the supervision of a classroom teacher. Only students’ first names were recorded for internal reference during administration; no personally identifiable information (e.g., full names, contact details, school IDs) was collected. For the purposes of analysis, all data were anonymized and used solely to examine aggregate patterns in student responses. The research protocol was designed to minimize risk and ensure confidentiality

%%%%%%%%%%%%%%%%%%%%%%%%%%%%%%%%%%%%%%%%%%%%%%%%%%%%%%%%%%%%%%%%%%%%%%%%

%%%%%%%%%%%%%%%%%%%%%%%%%%%%%%%%%%%%%%%%%%%%%%%%%%%%%%%%%%%%%%%%%%%%%%%%

%%% Use this environment to include acknowledgements (optional).
%%% This will be omitted in doubleblind mode.

\begin{ack}
We are grateful to the administration, faculty, and staff of Kendriya Vidyalaya, Indian Institute of Science, Bengaluru, for their institutional support and cooperation in facilitating this research. We sincerely thank all students, whose participation was essential for the successful completion of this study.
\end{ack}

%%%%%%%%%%%%%%%%%%%%%%%%%%%%%%%%%%%%%%%%%%%%%%%%%%%%%%%%%%%%%%%%%%%%%%%%

%%% Use this command to include your bibliography file.

\bibliography{ecai_preprint}

\begin{thebibliography}{47}
\providecommand{\natexlab}[1]{#1}
\providecommand{\url}[1]{\texttt{#1}}
\expandafter\ifx\csname urlstyle\endcsname\relax
  \providecommand{\doi}[1]{doi: #1}\else
  \providecommand{\doi}{doi: \begingroup \urlstyle{rm}\Url}\fi

\bibitem[Alhazmi et~al.(2024)Alhazmi, Sheng, Zhang, Zaib, and Alhazmi]{alhazmi2024distractor}
E.~Alhazmi, Q.~Sheng, W.~E. Zhang, M.~Zaib, and A.~Alhazmi.
\newblock Distractor generation in multiple-choice tasks: A survey of methods, datasets, and evaluation.
\newblock In \emph{Proceedings of the 2024 EMNLP}, pages 14437--14458, 2024.

\bibitem[Amidei et~al.(2018)Amidei, Piwek, and Willis]{amidei2018rethinking}
J.~Amidei, P.~Piwek, and A.~Willis.
\newblock Rethinking the agreement in human evaluation tasks.
\newblock In \emph{Proceedings of the 27th International COLING}, pages 3318--3329, 2018.

\bibitem[Anderson and Krathwohl(2001)]{blooms}
L.~W. Anderson and D.~R. Krathwohl.
\newblock \emph{A taxonomy for learning, teaching, and assessing: A revision of Bloom's taxonomy of educational objectives: complete edition}.
\newblock Addison Wesley Longman, Inc., 2001.

\bibitem[Bahdanau et~al.(2015)Bahdanau, Cho, and Bengio]{BahdanauCB14}
D.~Bahdanau, K.~Cho, and Y.~Bengio.
\newblock Neural machine translation by jointly learning to align and translate.
\newblock In \emph{{ICLR} 2015, San Diego, CA, USA, May 7-9, 2015, Conference Track Proceedings}, 2015.

\bibitem[Bai et~al.(2023)Bai, Bai, Chu, Cui, Dang, Deng, Fan, Ge, Han, Huang, et~al.]{bai2023qwen}
J.~Bai, S.~Bai, Y.~Chu, Z.~Cui, K.~Dang, X.~Deng, Y.~Fan, W.~Ge, Y.~Han, F.~Huang, et~al.
\newblock Qwen technical report.
\newblock \emph{arXiv preprint arXiv:2309.16609}, 2023.

\bibitem[Bjork et~al.(2014)Bjork, Little, and Storm]{bjork2014multiple}
E.~L. Bjork, J.~L. Little, and B.~C. Storm.
\newblock Multiple-choice testing as a desirable difficulty in the classroom.
\newblock \emph{Journal of Applied Research in Memory and Cognition}, 3\penalty0 (3):\penalty0 165--170, 2014.

\bibitem[Bojanowski et~al.(2017)Bojanowski, Grave, Joulin, and Mikolov]{bojanowski2017enriching}
P.~Bojanowski, E.~Grave, A.~Joulin, and T.~Mikolov.
\newblock Enriching word vectors with subword information.
\newblock \emph{TACL}, 5:\penalty0 135--146, 2017.

\bibitem[Butler(2018)]{BUTLER2018323}
A.~C. Butler.
\newblock Multiple-choice testing in education: Are the best practices for assessment also good for learning?
\newblock \emph{Journal of Applied Research in Memory and Cognition}, 7\penalty0 (3):\penalty0 323--331, 2018.
\newblock ISSN 2211-3681.

\bibitem[Cantor et~al.(2015)Cantor, Eslick, Marsh, Bjork, and Bjork]{cantor2015multiple}
A.~D. Cantor, A.~N. Eslick, E.~J. Marsh, R.~A. Bjork, and E.~L. Bjork.
\newblock Multiple-choice tests stabilize access to marginal knowledge.
\newblock \emph{Memory \& Cognition}, 43:\penalty0 193--205, 2015.

\bibitem[Cohen(1968)]{cohen1968weighted}
J.~Cohen.
\newblock Weighted kappa: nominal scale agreement provision for scaled disagreement or partial credit.
\newblock \emph{Psychological bulletin}, 70\penalty0 (4):\penalty0 213, 1968.

\bibitem[Collignon et~al.(2020)Collignon, Chacko, and Wydick~Martin]{collignon2020alternative}
S.~E. Collignon, J.~Chacko, and M.~Wydick~Martin.
\newblock An alternative multiple-choice question format to guide feedback using student self-assessment of knowledge.
\newblock \emph{Decision Sciences Journal of Innovative Education}, 18\penalty0 (3):\penalty0 456--480, 2020.

\bibitem[Das et~al.(2021)Das, Majumder, Phadikar, and Sekh]{das2021automatic}
B.~Das, M.~Majumder, S.~Phadikar, and A.~A. Sekh.
\newblock Automatic question generation and answer assessment: a survey.
\newblock \emph{Research and Practice in Technology Enhanced Learning}, 16\penalty0 (1):\penalty0 5, 2021.

\bibitem[Devlin et~al.(2019)Devlin, Chang, Lee, and Toutanova]{devlin-etal-2019-bert}
J.~Devlin, M.-W. Chang, K.~Lee, and K.~Toutanova.
\newblock {BERT}: Pre-training of deep bidirectional transformers for language understanding.
\newblock In \emph{Proceedings of the 2019 NAACL)}, pages 4171--4186, Minneapolis, Minnesota, June 2019. ACL.

\bibitem[Doughty et~al.(2024)Doughty, Wan, Bompelli, Qayum, Wang, Zhang, Zheng, Doyle, Sridhar, Agarwal, et~al.]{doughty2024comparative}
J.~Doughty, Z.~Wan, A.~Bompelli, J.~Qayum, T.~Wang, J.~Zhang, Y.~Zheng, A.~Doyle, P.~Sridhar, A.~Agarwal, et~al.
\newblock A comparative study of ai-generated (gpt-4) and human-crafted mcqs in programming education.
\newblock In \emph{Proceedings of the 26th Australasian Computing Education Conference}, pages 114--123, 2024.

\bibitem[Dubey et~al.(2024)Dubey, Jauhri, Pandey, Kadian, Al-Dahle, Letman, Mathur, Schelten, Yang, Fan, et~al.]{dubey2024llama}
A.~Dubey, A.~Jauhri, A.~Pandey, A.~Kadian, A.~Al-Dahle, A.~Letman, A.~Mathur, A.~Schelten, A.~Yang, A.~Fan, et~al.
\newblock The llama 3 herd of models.
\newblock \emph{arXiv preprint arXiv:2407.21783}, 2024.

\bibitem[Gallegos et~al.(2024)Gallegos, Rossi, Barrow, Tanjim, Kim, Dernoncourt, Yu, Zhang, and Ahmed]{gallegos-etal-2024-bias}
I.~O. Gallegos, R.~A. Rossi, J.~Barrow, M.~M. Tanjim, S.~Kim, F.~Dernoncourt, T.~Yu, R.~Zhang, and N.~K. Ahmed.
\newblock Bias and fairness in large language models: A survey.
\newblock \emph{Computational Linguistics}, 50\penalty0 (3):\penalty0 1097--1179, Sept. 2024.

\bibitem[Gao et~al.(2019)Gao, Bing, Li, King, and Lyu]{gao2019generating}
Y.~Gao, L.~Bing, P.~Li, I.~King, and M.~R. Lyu.
\newblock Generating distractors for reading comprehension questions from real examinations.
\newblock \emph{Proceedings of the AAAI Conference on Artificial Intelligence}, 33\penalty0 (01):\penalty0 6423--6430, 2019.

\bibitem[Gierl et~al.(2017)Gierl, Bulut, Guo, and Zhang]{gierl2017developing}
M.~J. Gierl, O.~Bulut, Q.~Guo, and X.~Zhang.
\newblock Developing, analyzing, and using distractors for multiple-choice tests in education: A comprehensive review.
\newblock \emph{Review of educational research}, 87\penalty0 (6):\penalty0 1082--1116, 2017.

\bibitem[Hadifar et~al.(2023)Hadifar, Bitew, Deleu, Develder, and Demeester]{hadifar2023eduqg}
A.~Hadifar, S.~K. Bitew, J.~Deleu, C.~Develder, and T.~Demeester.
\newblock Eduqg: A multi-format multiple-choice dataset for the educational domain.
\newblock \emph{IEEE Access}, 11:\penalty0 20885--20896, 2023.

\bibitem[Hurst et~al.(2024)Hurst, Lerer, Goucher, Perelman, Ramesh, Clark, Ostrow, Welihinda, Hayes, Radford, et~al.]{hurst2024gpt}
A.~Hurst, A.~Lerer, A.~P. Goucher, A.~Perelman, A.~Ramesh, A.~Clark, A.~Ostrow, A.~Welihinda, A.~Hayes, A.~Radford, et~al.
\newblock Gpt-4o system card.
\newblock \emph{arXiv preprint arXiv:2410.21276}, 2024.

\bibitem[Lewis et~al.(2020)Lewis, Liu, Goyal, Ghazvininejad, Mohamed, Levy, Stoyanov, and Zettlemoyer]{lewis-etal-2020-bart}
M.~Lewis, Y.~Liu, N.~Goyal, M.~Ghazvininejad, A.~Mohamed, O.~Levy, V.~Stoyanov, and L.~Zettlemoyer.
\newblock {BART}: Denoising sequence-to-sequence pre-training for natural language generation, translation, and comprehension.
\newblock In \emph{Proceedings of the 58th ACL}, pages 7871--7880, Online, July 2020. ACL.

\bibitem[Li et~al.(2015)Li, Luong, and Jurafsky]{li-etal-2015-hierarchical}
J.~Li, T.~Luong, and D.~Jurafsky.
\newblock A hierarchical neural autoencoder for paragraphs and documents.
\newblock In \emph{Proceedings of the 53rd ACL and the 7th IJNLP (Volume 1: Long Papers)}, pages 1106--1115, Beijing, China, July 2015. Association for Computational Linguistics.

\bibitem[Madri and Meruva(2023)]{madri2023comprehensive}
V.~R. Madri and S.~Meruva.
\newblock A comprehensive review on mcq generation from text.
\newblock \emph{Multimedia Tools and Applications}, 82\penalty0 (25):\penalty0 39415--39434, 2023.

\bibitem[Maity et~al.(2024)Maity, Deroy, and Sarkar]{maity2024novel}
S.~Maity, A.~Deroy, and S.~Sarkar.
\newblock A novel multi-stage prompting approach for language agnostic mcq generation using gpt.
\newblock In \emph{European Conference on Information Retrieval}, pages 268--277. Springer, 2024.

\bibitem[McDermott et~al.(2014)McDermott, Agarwal, D'Antonio, Roediger~III, and McDaniel]{mcdermott2014both}
K.~B. McDermott, P.~K. Agarwal, L.~D'Antonio, H.~L. Roediger~III, and M.~A. McDaniel.
\newblock Both multiple-choice and short-answer quizzes enhance later exam performance in middle and high school classes.
\newblock \emph{Journal of Experimental Psychology: Applied}, 20\penalty0 (1):\penalty0 3, 2014.

\bibitem[McHugh(2012)]{mchugh2012interrater}
M.~L. McHugh.
\newblock Interrater reliability: the kappa statistic.
\newblock \emph{Biochemia medica}, 22\penalty0 (3):\penalty0 276--282, 2012.

\bibitem[McNichols et~al.(2023)McNichols, Feng, Lee, Scarlatos, Smith, Woodhead, and Lan]{mcnichols2023exploring}
H.~McNichols, W.~Feng, J.~Lee, A.~Scarlatos, D.~Smith, S.~Woodhead, and A.~S. Lan.
\newblock Exploring automated distractor and feedback generation for math multiple-choice questions via in-context learning.
\newblock \emph{CoRR}, 2023.

\bibitem[Mikolov et~al.(2013)Mikolov, Sutskever, Chen, Corrado, and Dean]{mikolov2013distributed}
T.~Mikolov, I.~Sutskever, K.~Chen, G.~S. Corrado, and J.~Dean.
\newblock Distributed representations of words and phrases and their compositionality.
\newblock \emph{NeurIPS}, 26, 2013.

\bibitem[Pennington et~al.(2014)Pennington, Socher, and Manning]{pennington2014glove}
J.~Pennington, R.~Socher, and C.~D. Manning.
\newblock Glove: Global vectors for word representation.
\newblock In \emph{Proceedings of the 2014 EMNLP}, pages 1532--1543, 2014.

\bibitem[Radford et~al.(2019)Radford, Wu, Child, Luan, Amodei, Sutskever, et~al.]{radford2019language}
A.~Radford, J.~Wu, R.~Child, D.~Luan, D.~Amodei, I.~Sutskever, et~al.
\newblock Language models are unsupervised multitask learners.
\newblock \emph{OpenAI blog}, 1\penalty0 (8):\penalty0 9, 2019.

\bibitem[Raffel et~al.(2020)Raffel, Shazeer, Roberts, Lee, Narang, Matena, Zhou, Li, and Liu]{raffel2020exploring}
C.~Raffel, N.~Shazeer, A.~Roberts, K.~Lee, S.~Narang, M.~Matena, Y.~Zhou, W.~Li, and P.~J. Liu.
\newblock Exploring the limits of transfer learning with a unified text-to-text transformer.
\newblock \emph{JMLR}, 21\penalty0 (140):\penalty0 1--67, 2020.

\bibitem[Reimers and Gurevych(2019)]{sentence}
N.~Reimers and I.~Gurevych.
\newblock Sentence-{BERT}: Sentence embeddings using {S}iamese {BERT}-networks.
\newblock In K.~Inui, J.~Jiang, V.~Ng, and X.~Wan, editors, \emph{Proceedings of the 2019 EMNLP-IJCNLP}, pages 3982--3992, Hong Kong, China, Nov. 2019. ACL.

\bibitem[Seo et~al.(2017)Seo, Kembhavi, Farhadi, and Hajishirzi]{seo2017bidirectional}
M.~Seo, A.~Kembhavi, A.~Farhadi, and H.~Hajishirzi.
\newblock Bidirectional attention flow for machine comprehension.
\newblock In \emph{ICLR}, 2017.

\bibitem[Shuai et~al.(2021)Shuai, Wei, Liu, Xu, and Li]{9533341}
P.~Shuai, Z.~Wei, S.~Liu, X.~Xu, and L.~Li.
\newblock Topic enhanced multi-head co-attention: Generating distractors for reading comprehension.
\newblock In \emph{2021 IJCNN}, pages 1--8, 2021.

\bibitem[Song et~al.(2020)Song, Tan, Qin, Lu, and Liu]{song2020mpnet}
K.~Song, X.~Tan, T.~Qin, J.~Lu, and T.-Y. Liu.
\newblock Mpnet: Masked and permuted pre-training for language understanding.
\newblock \emph{NeurIPS}, 33:\penalty0 16857--16867, 2020.

\bibitem[States(2013)]{NGSS}
N.~L. States.
\newblock \emph{Next generation science standards: For states, by states}.
\newblock National Academies Press, 2013.

\bibitem[Sutskever et~al.(2014)Sutskever, Vinyals, and Le]{ilya}
I.~Sutskever, O.~Vinyals, and Q.~V. Le.
\newblock Sequence to sequence learning with neural networks.
\newblock In \emph{Proceedings of the 28th NIPS - Volume 2}, NIPS'14, page 3104–3112, Cambridge, MA, USA, 2014. MIT Press.

\bibitem[Team et~al.(2024)Team, Georgiev, Lei, Burnell, Bai, Gulati, Tanzer, Vincent, Pan, Wang, et~al.]{team2024gemini}
G.~Team, P.~Georgiev, V.~I. Lei, R.~Burnell, L.~Bai, A.~Gulati, G.~Tanzer, D.~Vincent, Z.~Pan, S.~Wang, et~al.
\newblock Gemini 1.5: Unlocking multimodal understanding across millions of tokens of context.
\newblock \emph{arXiv preprint arXiv:2403.05530}, 2024.

\bibitem[Thomas et~al.(2024)Thomas, Borchers, Kakarla, Lin, Bhushan, Guo, Gatz, and Koedinger]{thomas2024does}
D.~R. Thomas, C.~Borchers, S.~Kakarla, J.~Lin, S.~Bhushan, B.~Guo, E.~Gatz, and K.~R. Koedinger.
\newblock Does multiple choice have a future in the age of generative ai? a posttest-only rct.
\newblock \emph{arXiv preprint arXiv:2412.10267}, 2024.

\bibitem[Urone and Hinrichs(2020)]{urone_hinrichs_2020}
P.~P. Urone and R.~Hinrichs.
\newblock \emph{Physics}.
\newblock OpenStax, Houston, Texas, 2020.
\newblock URL \url{https://openstax.org/books/physics}.

\bibitem[Wang et~al.(2023)Wang, Bai, Rong, Ouyang, and Xiong]{wang2023weak}
J.~Wang, J.~Bai, W.~Rong, Y.~Ouyang, and Z.~Xiong.
\newblock Weak positive sampling and soft smooth labeling for distractor generation data augmentation.
\newblock In \emph{International Conference on Intelligent Computing}, pages 756--767. Springer, 2023.

\bibitem[Wei et~al.(2022)Wei, Wang, Schuurmans, Bosma, Xia, Chi, Le, Zhou, et~al.]{wei2022chain}
J.~Wei, X.~Wang, D.~Schuurmans, M.~Bosma, F.~Xia, E.~Chi, Q.~V. Le, D.~Zhou, et~al.
\newblock Chain-of-thought prompting elicits reasoning in large language models.
\newblock \emph{NeurIPS}, 35:\penalty0 24824--24837, 2022.

\bibitem[Welbl et~al.(2017)Welbl, Liu, and Gardner]{welbl-etal-2017-crowdsourcing}
J.~Welbl, N.~F. Liu, and M.~Gardner.
\newblock Crowdsourcing multiple choice science questions.
\newblock In \emph{Proceedings of the 3rd Workshop on Noisy User-generated Text}, pages 94--106. ACL, Sept. 2017.

\bibitem[Zellers et~al.(2018)Zellers, Bisk, Schwartz, and Choi]{zellers-etal-2018-swag}
R.~Zellers, Y.~Bisk, R.~Schwartz, and Y.~Choi.
\newblock {SWAG}: A large-scale adversarial dataset for grounded commonsense inference.
\newblock In \emph{Proceedings of the 2018 EMNLP}, pages 93--104. ACL, Oct.-Nov. 2018.

\bibitem[Zhang et~al.(2023)Zhang, Li, Cui, Cai, Liu, Fu, Huang, Zhao, Zhang, Chen, et~al.]{zhang2023siren}
Y.~Zhang, Y.~Li, L.~Cui, D.~Cai, L.~Liu, T.~Fu, X.~Huang, E.~Zhao, Y.~Zhang, Y.~Chen, et~al.
\newblock Siren's song in the ai ocean: a survey on hallucination in large language models.
\newblock \emph{arXiv preprint arXiv:2309.01219}, 2023.

\bibitem[Zheng et~al.(2024)Zheng, Chiang, Sheng, Zhuang, Wu, Zhuang, Lin, Li, Li, Xing, et~al.]{zheng2024judging}
L.~Zheng, W.-L. Chiang, Y.~Sheng, S.~Zhuang, Z.~Wu, Y.~Zhuang, Z.~Lin, Z.~Li, D.~Li, E.~Xing, et~al.
\newblock Judging llm-as-a-judge with mt-bench and chatbot arena.
\newblock \emph{NeurIPS}, 36, 2024.

\bibitem[Zhou et~al.(2020)Zhou, Luo, and Wu]{zhou2020co}
X.~Zhou, S.~Luo, and Y.~Wu.
\newblock Co-attention hierarchical network: Generating coherent long distractors for reading comprehension.
\newblock \emph{Proceedings of the AAAI Conference on Artificial Intelligence}, 34\penalty0 (05):\penalty0 9725--9732, 2020.

\end{thebibliography}

% \clearpage
% \appendix

% \section{Appendix}

\end{document}